\documentclass{article}
\usepackage[utf8]{inputenc}
\usepackage[english]{babel}
\usepackage{amsmath,amsfonts,amssymb}
\usepackage{amsthm}
\usepackage{mathtools}
\usepackage[margin=1in]{geometry}
\usepackage{graphicx}
\usepackage{subcaption}
\usepackage{tikz}
\usepackage{booktabs}
\usepackage{longtable}
\usepackage{array}
\usepackage{adjustbox}
\usepackage[numbers,sort&compress]{natbib}
\usepackage{url}
\usepackage[breaklinks=true,colorlinks=true,citecolor=blue,linkcolor=blue,urlcolor=blue]{hyperref}

\usepackage{enumerate}
\usepackage{algorithm}
\usepackage{algorithmic}
\usepackage{xcolor}

\theoremstyle{plain}

\theoremstyle{definition}

\title{Developing an AI-Guided Assistant Device for the Deaf and Hearing Impaired}

\author{Jiayu (Jerry) Liu}
\date{\today}

\begin{document}

\maketitle

 \begin{abstract}
\textbf{Purpose:} This study aims to develop a deep learning system for an accessibility device for the deaf or hearing impaired. The device will accurately localize and identify sound sources in real time. This study will fill an important gap in current research by leveraging machine learning techniques to target the underprivileged community. 

\textbf{Design and Methods:} The system includes three main components. 

\begin{itemize}
    \item JerryNet: A custom designed CNN architecture that determines the direction of arrival (DoA) for nine possible directions. The input is in the form of a phase matrix generated by the concurrent audio of four different microphones.  
    \item Audio Classification: This model is based on fine-tuning the Contrastive Language-Audio Pretraining (CLAP) model to identify the exact sound classes only based on audio. 
    \item Multimodal integration model: This is an accurate sound localization model that combines audio, visual, and text data to locate the exact sound sources in the images. The part consists of two modules, one object detection using Yolov9 to generate all the bounding boxes of the objects, and an audio visual localization model to identify the optimal bounding box using complete Intersection over Union (CIoU). 
\end{itemize}

The hardware consists of a four-microphone rectangular formation and a camera mounted on glasses with a wristband for displaying necessary information like direction. 

\textbf{Results:} On a custom collected data set, JerryNet achieved a precision of 91. 1\% for the sound direction, outperforming all the baseline models. The CLAP model achieved 98.5\% and 95\% accuracy on custom and AudioSet datasets, respectively. The audio-visual localization model within component 3 yielded a cIoU of $0.892$ and an AUC of $0.658$, surpassing other similar models. 

\textbf{Conclusion:} This work advances the capability of assistive technologies particularly for the deaf and hearing impaired. There are many future potentials to this study, paving the way to creating a new generation of accessibility devices.  

\end{abstract}  

\section{Introduction}
\label{sec:introduction}

\textbf{Background and Context: } This project aims to produce an assistant device for people with deaf or hearing impairment that is more accessible to a wider spectrum of financial status. Born with congenital hearing loss certainly presents many difficulties; however, thanks to my family, I have been gifted with the ability to hear through the technology of a cochlear implant. This device is quite expensive, with a price of around 50000 USD for both left and right ears. Although the calculation of this cost includes surgeries, hospitalization, and the majority of accessories or procedures required to maintain device health many years after use, the staggering number stood as a barricade for deaf people to survive in a rapidly changing society. In contrast, the combined cost for raw components and manufacturing for my device barely exceeds 20 USD, something attainable by essentially all deaf or hearing loss populations. Through the cooperation of multiple microphones and cameras, my device offers a multimodal approach to assistant technologies, which can simultaneously perform various tasks, a unique method for assistant devices with fewer prior research. My device includes two constituent parts: a pair of glasses equipped with microphones and a camera for audio and visual input, and a wristband for user display. The multitude of functions of this device includes sound localization, audio classification, emergency precaution, and finally text-to-speech. A smooth user experience is established through a versatile amalgamation of all these functions into one control loop, providing a more reliable environment for deaf or hearing impaired people to rely on in society. A more detailed explanation of all these functions and the main control loop will be discussed in depth in future sections. Finally, I believe that the steps of my software components can be extracted and modified to meet the needs of other people, laying the foundation for a broader ecosystem of assistant technologies that improve quality of life. 

\textbf{Research Question:} How to identify the exact location of a sound source through a combination of audio and visual input and also classify the sound-making object using audio effectively for application in an assistant device?

\textbf{Thesis Statement:} Through the design and application of several deep learning models, I can fulfill the fundamental task of sound localization and audio classification; this will not only bridge the missing gap between frontier research in artificial intelligence with assistant devices for the auditory impaired, but also explore new potential fields that provide a foundation for many novel scientific innovations and societal applications that can extend a helping hand to all kinds of people. 

\textbf{Paper Outline:} This paper will first describe in detail the system architecture of the assistant device and provide an explanation for each deep learning model implemented in each step. Then, the paper will discuss the evaluation metric of the different models and conclude whether the training results actually achieve the engineering goals and answer the research questions. 

\section{Method}
\label{sec:method}

\subsection{Model Architectures}

\subsection{Overview of System Architecture and Hardware Implementation}

\begin{enumerate}
    \item Identify DoA (Direction of Arrival) with only audio input: The first component will receive an audio clip and is responsible for determining the direction of the arrival of the sound. There will be nine possible directions of arrival, with eight representing the different direction and one representing the person wearing the device speaking. 
    \item Zero-shot audio classification: The second component is a model that will also receive a singular audio clip and is assigned the task of labeling the audio with some pre-defined set of sound-producing objects. 
    \item Sound Localization Through Combination of Visual, Auditory, and Textual Input: The final component will seek to combine and integrate all available information for a high-accuracy sound localization. This model will receive input from the DoA designated in component $1$ and the audio class from component $2$. Then, finally, using an image, the model will identify the exact object in the picture (even if the same type of object appears multiple times) and place a bounding box around it for display. 
\end{enumerate}

\begin{figure}
    \centering
    \includegraphics[width=0.7\linewidth]{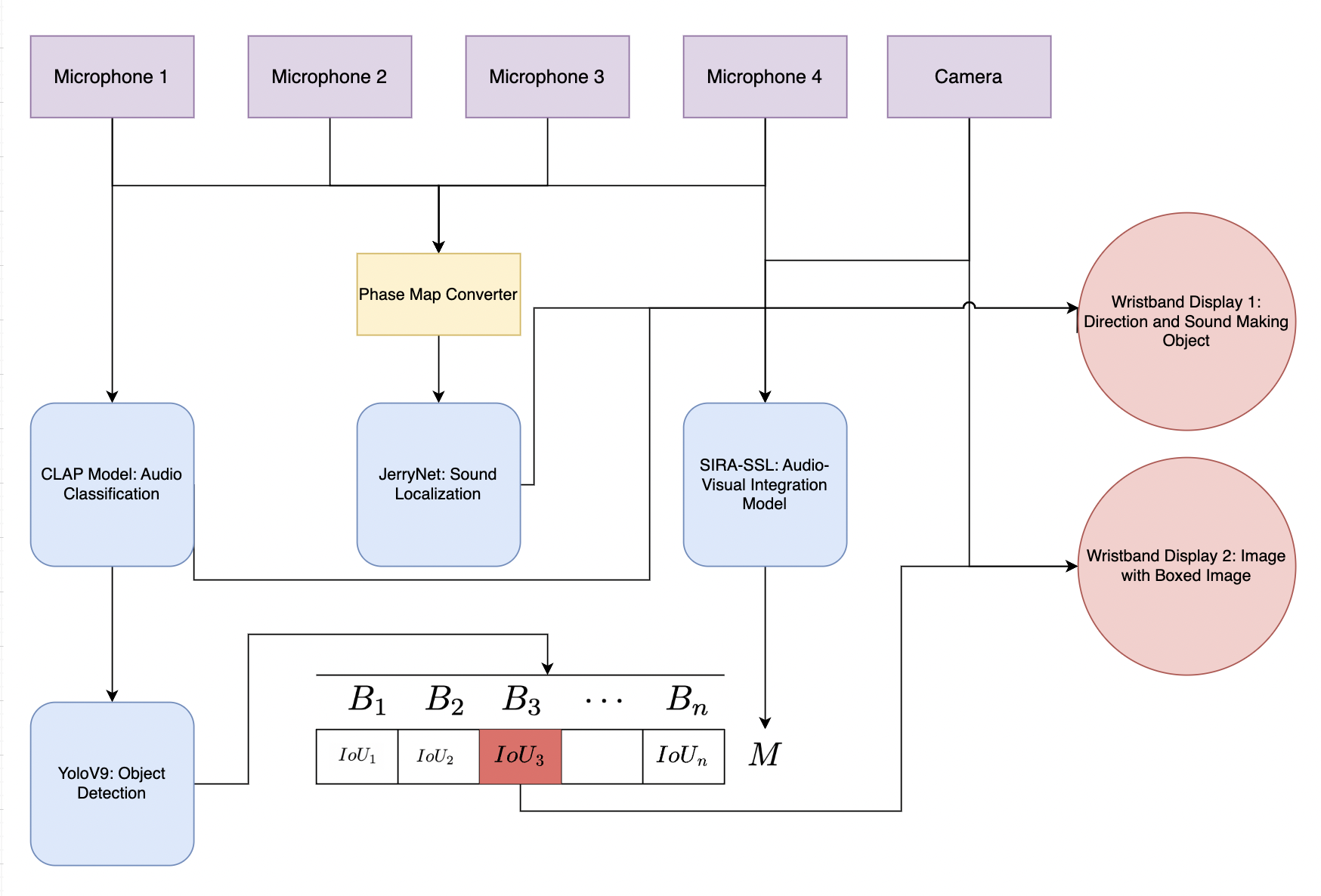}
    \caption{Flowchart of system architecture}
\end{figure}

\textbf{Hardware Design: } The hardware components include glasses equipped with four microphones and a camera and a watch with a display screen. The hardware will execute the system architecture as follows: the four microphones on the glasses will repeatedly send a $2$ second audio recording to the central server. The central server processes the audio segment, including checking if the decibel reaches $60$, a reference to eliminate insignificant sounds. If so, the central server passes the audio signal to both the model in component $1$ and component $2$ for evaluation, determining the Direction of Arrival and the audio classes. The wristband will then display the following. There is an 'object' in the 'direction' where the object and direction are outputs of the model. Afterward, the person is expected to turn in that direction, which will signal the camera on the glasses to take a photo, which is sent to the central server. Finally, component 3, with DoA and audio classes information, will be used to generate a bounding box surrounding the object that produced the sound, which will be displayed on the wristband. This is a complete cycle of the assistant device, and a new cycle will not commence if the previous cycle is still running. 

\begin{figure}[h]
    \centering
    \begin{subfigure}{0.45\linewidth}
        \centering
        \includegraphics[width=\linewidth]{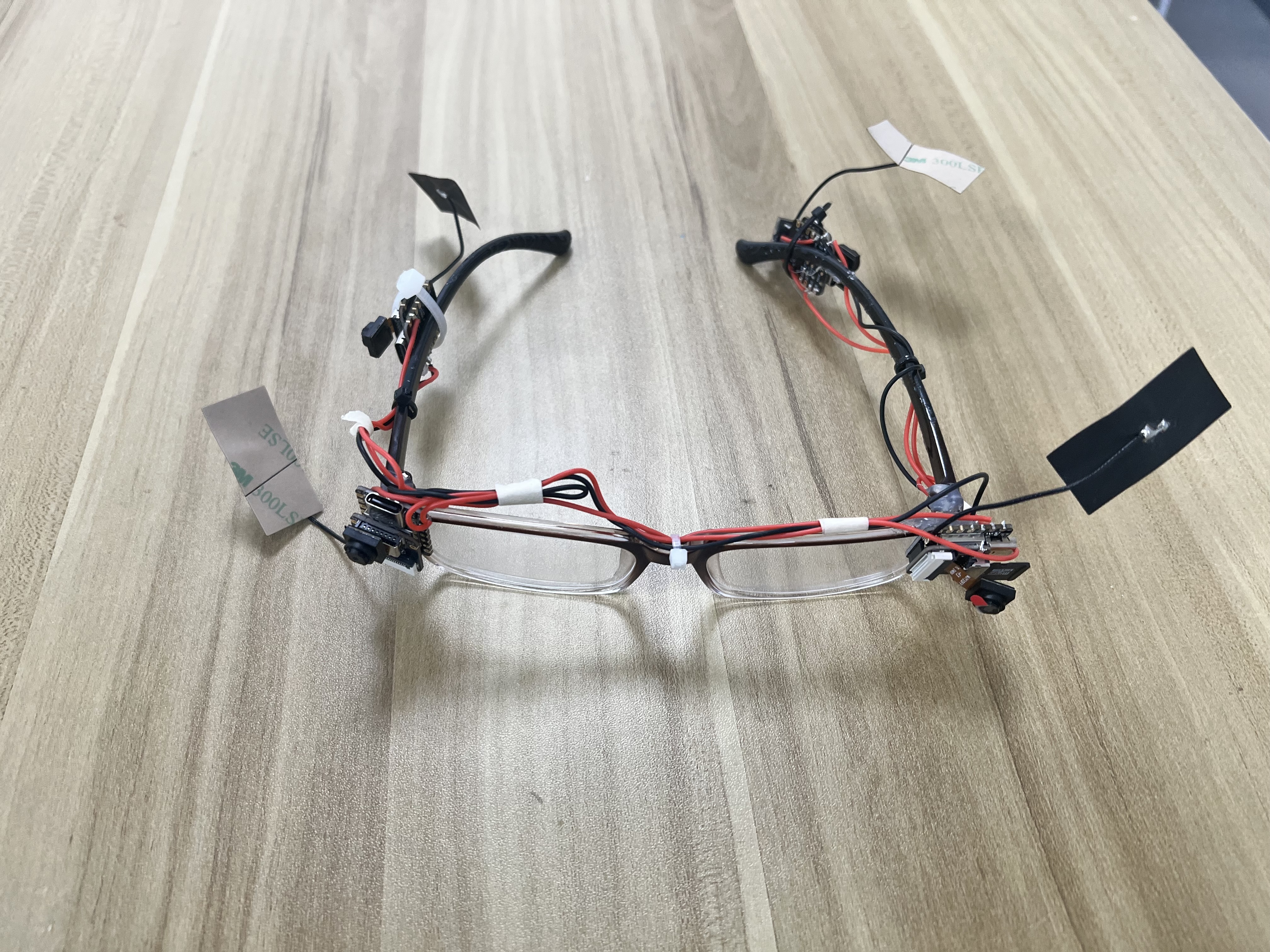}
        \label{fig:glasses}
    \end{subfigure}
    \hfill
    \begin{subfigure}{0.45\linewidth}
        \centering
        \includegraphics[width=\linewidth]{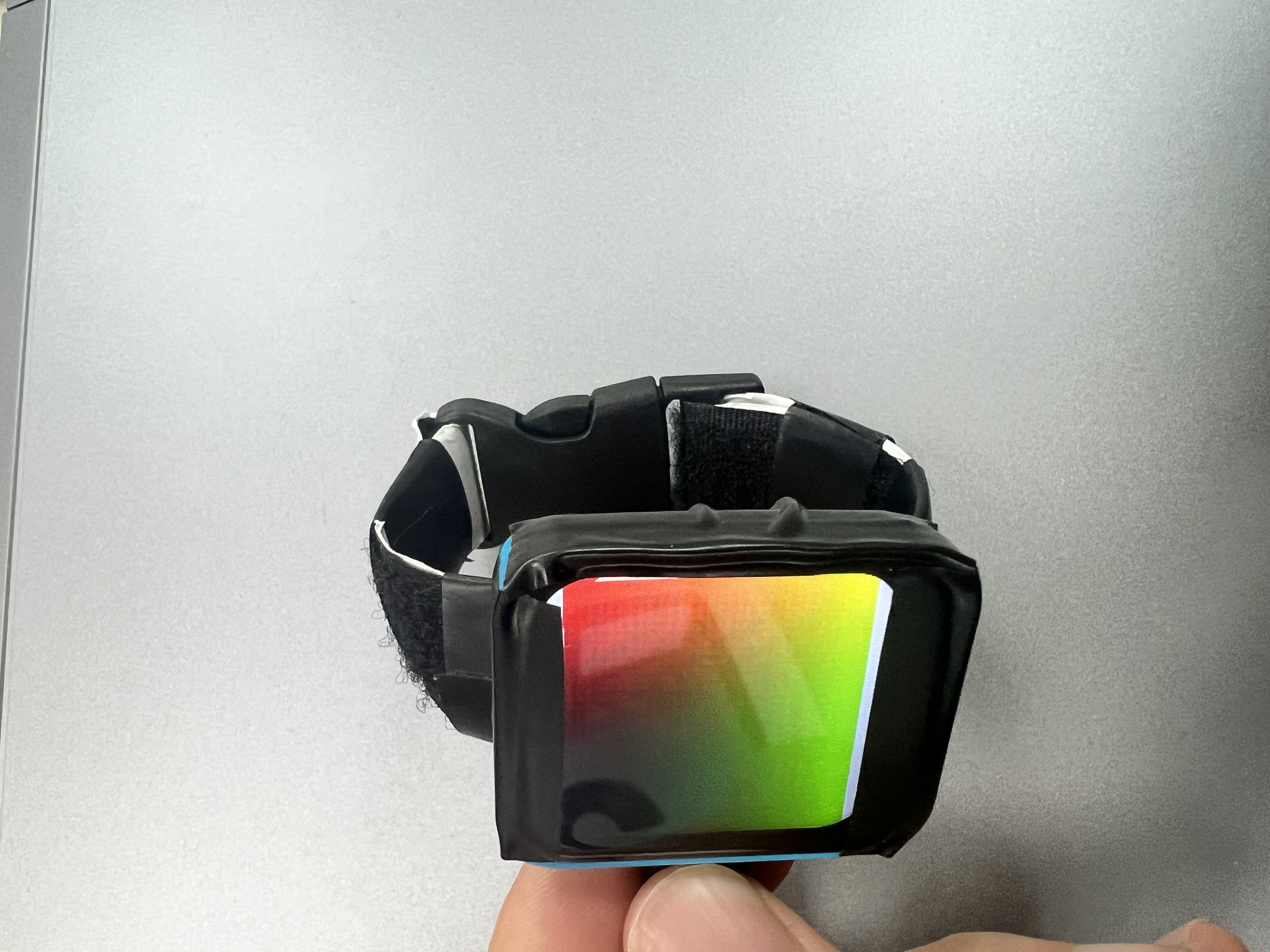}
        \label{fig:watch}
    \end{subfigure}
    \caption{Photo of homemade glasses and wristband with display}
    \label{fig:audio_visual}
\end{figure}

\subsection{Model 1: Sound Localization via Audio}

The first component of our system aims to identify the location of the sound producing object using only audio information. Instead of a classical localization method, such as triangulation, we strive for a more sophisticated machine learning approach. More specifically, our device incorporates four microphones arranged in a fixed rectangular configuration attached to the glasses as seen in Figure 1 where each of them records audio continuously and concurrently. We then exploit the small difference between the four audio channels to determine the direction of the sound. We adopt a custom CNN model, referred to as \textbf{JerryNet}. As benchmarks, we trained JerryNet alongside well-known architectures like \textbf{VGGNet}, \textbf{ResNet50}, \textbf{DenseNet121}, and \textbf{BAST} (Binaural Audio Spectrogram Transformer).

\subsubsection{Input Conversion: Phase Map}

Firstly, to extract spatial information from raw audio, we will convert the raw audio into a time-frequency representation. Each microphone’s signal is transformed using the Short-Time Fourier Transform (STFT):
\[ X_i(f,\tau) = \sum_{n=0}^{N-1} x_i(\tau \cdot H + n)\, w(n) \, e^{-j 2\pi \frac{n f}{N}}, \]
where \(x_i(t)\) denotes the audio signal from the \(i\)-th microphone, \(w(n)\) is a window function, \(H\) is the hop size, and \(f\) and \(\tau\) index frequency bins and time frames, respectively.

Since each STFT coefficient \(X_i(f,\tau)\) is complex-valued, it contains both magnitude and phase information. The differences in phases between the microphones is critical for revealing the sound directions. Therefore, using this, we will define the interchannel phase difference (IPD) between two microphones \(i\) and \(j\) as 

\[ \text{IPD}_{ij}(f,\tau) = \angle X_i(f,\tau) - \angle X_j(f,\tau), \]

with \(\angle (\cdot)\) denoting the phase angle.

Since we have a four-microphone configuration, we can designate the first microphone as the reference and calculate 
\[ \text{IPD}_{1j}(f,\tau) = \angle X_1(f,\tau) - \angle X_j(f,\tau), \quad j = 2,3,4. \]

The IPDS will be assembled into a \emph{phase matrix}
\[
    \Phi(f,\tau) =
    \begin{bmatrix}
        \text{IPD}_{12}(f,\tau) \\
        \text{IPD}_{13}(f,\tau) \\
        \text{IPD}_{14}(f,\tau)
    \end{bmatrix}.
\]

This phase matrix \(\Phi(f,\tau)\) will serve as the input for the JerryNet due to its ability to preserve the IPD necessary for accurately estimating the Direction of Arrival (DoA). 

\subsubsection{JerryNet: A CNN-based Architecture for Sound Localization}

JerryNet is a custom designed CNN model to accomplish sound localization using input as phase matrix. It takes one-channel input as the phase matrix and employs multiple convolutional layers with small kernels, size \(1 \times 2\) with unit strides to model the fine-grained temporal variation. The model consists of ten blocks, where each block consists of multiple convolutional layers, followed by ReLU activation, and finally ending with one MaxPooling layer. 

The respective feature maps become flattened upon extraction through the convolution process and further feed into the following series of fully connected layers. Concretely, a first fully connected layer projects the flattened features to a high-dimensional space of 4096 units, which is further refined into an intermediate layer of 256 units. Finally, a final linear layer outputs raw logits corresponding to the 9 possible directional classes. 

The networks are trained using standard backpropagation and mini-batch gradient descent. Given our relatively small dataset, heavy data augmentation and careful hyperparameter tuning (e.g., for learning rate, batch size) are employed to mitigate overfitting and enhance model robustness.

We adopt a \textbf{Binary Cross-Entropy (BCE)} loss formulation to treat each of the 9 possible directions as an independent class. Internally, the network outputs a 9-dimensional softmax distribution, and the highest-probability direction is taken as the final DoA prediction.

\begin{figure}[h]
    \centering
    \includegraphics[width=0.6\linewidth]{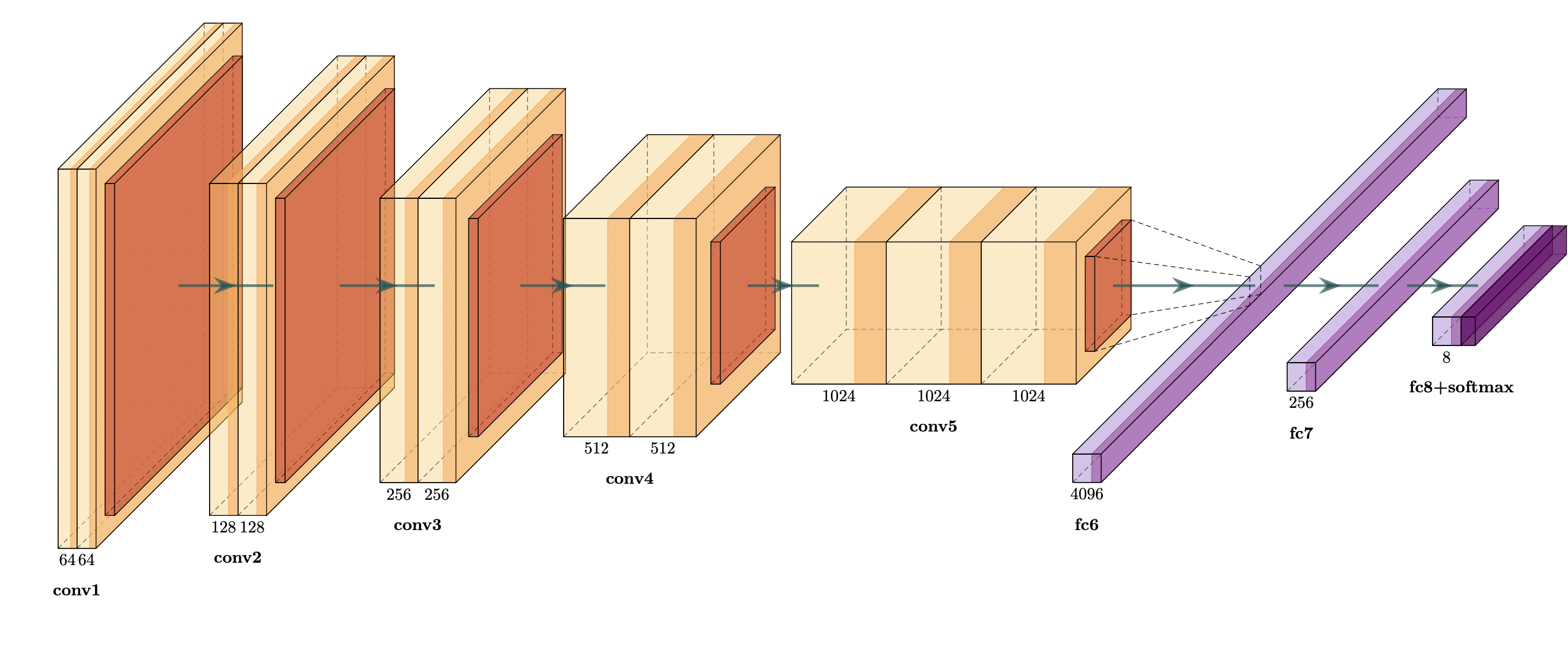}
    \caption{Schematic of the CNN Architecture for Sound Localization (JerryNet)}
    \label{fig:CNN_architecture}
\end{figure}

\subsubsection{Data Collection}

We have gathered a custom dataset consisting of exactly $450$ sets of four concurrent microphone audio data, where there are exactly $50$ sets for each of the nine distinct directions of arrival. The data is allocated between indoor and outdoor environments to increase model accuracy for different background environments. In fact, many of the data are intentionally designed such that the model can be trained to essentially isolate important sounds even in noisy conditions in outdoor or indoor settings with people talking. 

For the data collection, I would be wearing the glasses with the four microphones and standing in a room or an outdoor environment. Then, I would have my friend or teacher produce a designated sound in each of the eight directions. The ninth direction is recorded by me, the user, speaking while wearing the glasses. The audio input from the four microphones is amplified to exaggerate the volume difference for a more significant result. 

\subsection{Model 2: Zero-shot Audio Classification}

The second component of the system architecture is dedicated to classifying the type of object that produces the sound purely from audio. We will leverage the \textbf{CLAP (Contrastive Language-Audio Pretraining)} model, which is inspired by OpenAI's CLIP \cite{elizalde2022clap}.

\subsubsection{CLAP Model Architecture}

\begin{figure}[h]
    \centering
    \includegraphics[width=0.6\linewidth]{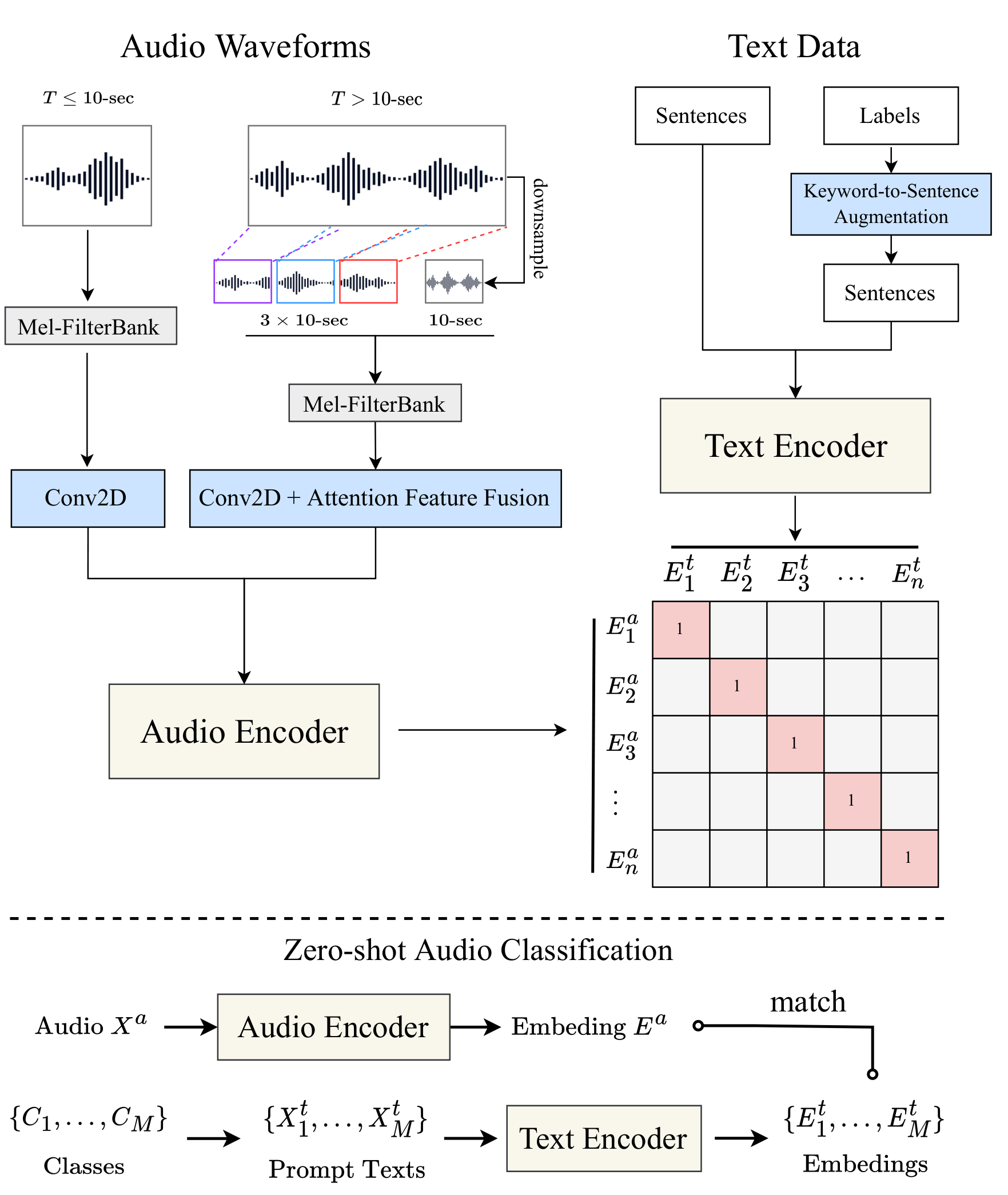}
    \caption{Diagram of CLAP Architecture \cite{elizalde2022clap}}
    \label{CLAP}
\end{figure}

The CLAP model employs a dual encoder, one audio encoder, and one text encoder. The model architecture is shown in Figure \ref{CLAP}. 

The audio encoder is a \textbf{SwinTransformer}, which captures important frequency information over time. SwinTransformers has the capability to capture both local and global information, particularly by enhancing performance on sequential audio data. This encoder allows the model to operate efficiently even in complex environments, even with overlapping conversations. 

The text encoder is a \textbf{RoBERTa}, which has shown prior superior performance in natural language processing tasks due to its pre-training in a large collection of written and spoken text, helping it associate textual labels with diverse audio. 

Once the embeddings are generated, an MLP layer with ReLU activation reduces the dimensionality to 512 to ensure alignment between the audio and textual spaces. 

The specific task we hope to achieve is \textbf{zero-shot audio classification}, which essentially matches an audio clip with a single prompt text from a list of text prompts representing all the different possible sound classes. This task is completed through generating cosine similarities between the audio embedding vector and all other text embedding vectors, which are passed through a softmax to generate a probability distribution over all the predefined sound classes. 

\subsubsection{Importance Classifier}

Firstly, the CLAP model will generate a probability score for each of the sound classes. Then, the important classifier is tasked with filtering the most relevant sound for the user. We will use a threshold of $0.3$ where all sounds with a probability score above $0.3$ will be passed into the classifier. This experimentation, where a multisound environment will typically result in $0.3$ for each sound. The classifier would refer to a priority list of sounds more relevant to a deaf or hearing-impaired person, in combination with their relative confidence scores. The priority list is shown below. 

\begin{table}[h!]
    \centering
    \begin{tabular}{|c|l|}
        \hline
        \textbf{Relative Priority} & \textbf{Class Name} \\ \hline
        1  & Siren   \\ \hline
        2  & Car Honking   \\ \hline
        3  & Bike Bell      \\ \hline
        4  & Person Talking \\ \hline
        5  & Doorbell         \\ \hline
        6  & Phone Ringing    \\ \hline
        7  & Dog Barking   \\ \hline
        8  & Instruments    \\ \hline
    \end{tabular}
    \caption{Examples of Sound Priority}
    \label{tab:output_classes}
\end{table}

\subsubsection{Data Collection and Fine-Tuning}

We have fine-tuned the CLAP model using the custom dataset from Part 1. The dataset will be divided into 80-20 splits for training and testing. This fine-tuning will enhance the CLAP model since it will train it on a dataset that mimics better real complex sound environments. The fine-tuning dataset is capable of high accuracy, identifying sound classes even in a chaotic environment. 

\subsection{Model 3: Multimodal Integration of Audio, Textual and Visual for Accurate Localization}

The third component involves the integration of audio, textual, and visual modes for accurate sound localization. This is the most effective because humans also naturally localize sound through the combination of these three modes. For example, when we were young, we developed the skill of corresponding audio sounds to specific objects within the field of vision, such as a human face with a sound of speech, a cow with the sound of ``moo'', etc. 

\subsubsection{Module 1: Object Detection}

The first module of the third component will generate multiple boundary boxes based on the textual input of the model $2$ and the images. For example, if model $2$ labeled the audio as a person, the object detection model will box all the persons that appear in the photo. We will use Yolov9 for fast and efficient image classification and object identification. Then, the multiple bounding boxes for the designated object will be input to the component $2$ below. 

The pretrained Yolov9 model only contains $80$ object classes, which do not include many of our designated sound-producing objects included in our fine-tuned CLAP model. Therefore, we will fine-tune Yolov9 using the Open Image Dataset from Google to include all the sound classes.  

\subsubsection{Module 2: Box Selection Algorithm}

Next, we need to utilize the box selection algorithm to isolate the exact object that is producing the sound. The box selection algorithm will select the optimal box from all the bounding boxes generated from the object detection model. This model will consider two main factors: the DoA received from part 1 and also a localization map generated through an audio-visual source localization model.

We will mainly use a metric called the complete Intersection over Union (cIoU), which essentially measures the area of intersection between the localization map and the several bounding boxes. 

Here, we outline the computation process for the index cIoU. 

\begin{enumerate}
    \item Thresholding the Localization Map: The continuous localization map output from the audio visual sound localization model is converted to a binary map by applying a threshold. Pixels above the threshold are considered as "positive" detections.
    \item Creating a Pseudo-Bounding Box: From the thresholded binary map, a bounding box is derived by finding the smallest rectangle that encompasses all positive pixels.
    \item Comparison with Ground Truth: The derived pseudo-bounding box is then compared with the ground truth bounding box using the standard IoU calculation, which is shown below. 
    \[ \text{IoU} = \frac{|B_p \cap B_{gt}|}{|B_p \cup B_{gt}|} \]
    where $B_gt$ is the ground truth bounding box of the audio visual localization model produced from the two above steps and $B_p$ is the bounding box computed from the image segmentation models. 
    \item Selection of the Maximum IoU Value: The bounding boxes that achieve the maximum IoU value with $B_{gt}$ will finally be selected. 
\end{enumerate}

\subsubsection{Audio-Visual Sound Localization Model: }

Finally, after extensive research on multiple generations of the audio visual sound localization model, including AVL, EZ-VSL, and LVS. I have decided to implement the SIRA-SSL, whose full name is Audio-Visual Spatial Integration and Recursive Attention for Robust Sound Source Localization. The model architecture is quite complicated; nevertheless, the crux consists of two components: audio-visual spatial integration network and a recursive attention network. The audio-visual spatial integration passes the image and audio through separate encoders, then combines the output through attention modules to produce two initial localization maps: a visual attentive and an audio attentive localization map. Then these localization maps and the original image are passed into a recursive attention network, which mimics how humans repeatedly focus on sound-making objects for a more accurate sound localization, effectively eliminating unnecessary regions. The model will be trained on the data provided from their Github \cite{audio_visual_integration}. 

\begin{figure}[h]
    \centering
    \includegraphics[width=0.3\linewidth]{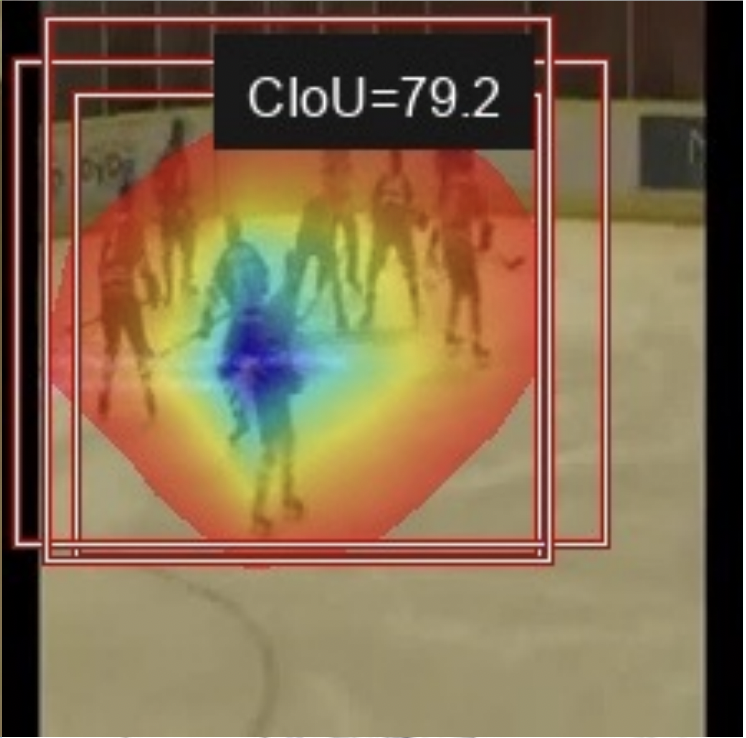}
    \caption{Image with Localization Map, Bounding Boxes, and cIoU value}
    \label{fig:enter-label}
\end{figure}

\section{Results}

\subsection{Model 1: Sound Localization Purely through Audio}

The first component aimed to determine the Direction of Arrival (DoA) using only audio inputs from four microphones. We compared our custom JerryNet model with several baseline architectures, including VGG16, VGG19, ResNet50V2, DenseNet121, and BAST. All models were fine-tuned on our custom dataset of 300 audio-direction pairs.

\subsubsection{Performance Comparison}

The models were evaluated using accuracy and F1 score as metrics, given the multiclass nature of the DoA classification task. Table \ref{tab:model1_performance} summarizes the results.

\begin{table}[h!]
    \centering
    \begin{tabular}{|l|c|c|}
        \hline
        \textbf{Model} & \textbf{Accuracy (\%)} & \textbf{F1-Score} \\ \hline
        JerryNet & \textbf{91.1} & \textbf{0.910} \\ \hline
        VGG16 & 83.3 & 0.832 \\ \hline
        VGG19 & 84.5 & 0.843 \\ \hline
        ResNet50V2 & 85.3 & 0.850 \\ \hline
        DenseNet121 & 87.0 & 0.867 \\ \hline
        BAST & 88.2 & 0.880 \\ \hline
    \end{tabular}
    \caption{Performance for Different CNN}
    \label{tab:model1_performance}
\end{table}

JerryNet outperformed all baseline models, achieving an accuracy of 91.1\% and an F1 score of 0.910. This implies that the model can successfully classify the direction of sound out of the nine possible options approximately $91$ percent of the time, which is slightly better compared with others. Here is the accuracy graph for the CNN model.

\begin{figure}[h]
    \centering
    \includegraphics[width=0.5\linewidth]{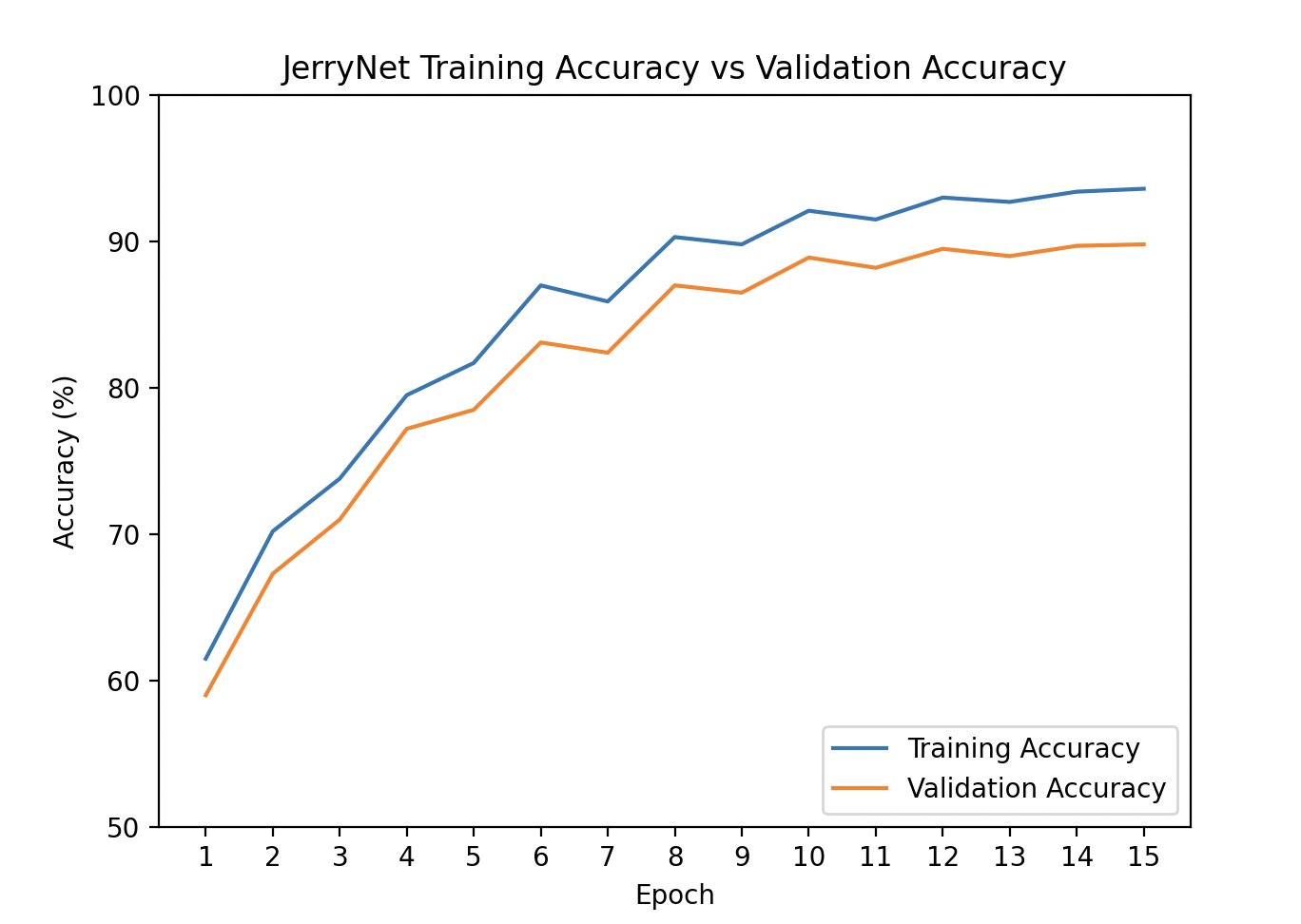}
    \caption{Training/Validation accuracy of JerryNet}
\end{figure}

In order to verify formally whether the JerryNet actually performs statistically significantly better than the other model, we will conduct a one-way ANOVA test followed by Tukey's HSD post hoc analysis. The ANOVA results indicated a significant difference among the models (\(F(5, 114) = 9.28\), \(p < 0.001\)). Tukey's HSD test confirmed that JerryNet's performance was significantly better than that of the baseline models (\(p < 0.05\)).

\subsection{Model 2: Zero-shot Audio Classification}

The fine-tuned CLAP model achieved an accuracy of \(98.5\%\) for the test set generated from my custom collected data set. To further verify its accuracy, we used approximately 5000 labeled audio clips from the AudioSet dataset, where the fine-tuned model achieved a \(95\%\) accuracy. 

\subsection{Model 3: Multimodal Integration for Accurate Localization}

We evaluated the audio-visual sound localization aspect of model 3 using the cIoU and AUC metrics on a test set of 100 images with corresponding audio clips. The results are shown in Table \ref{tab:multimodal_performance}.

\begin{table}[h!]
    \centering
    \begin{tabular}{|l|c|c|}
        \hline
        \textbf{Metric} & \textbf{Our System} & \textbf{Baseline (LVS)} \\ \hline
        cIoU & \textbf{0.892} & 0.876 \\ \hline
        AUC & \textbf{0.658} & 0.641 \\ \hline
    \end{tabular}
    \caption{Performance of Multimodal Integration Model}
    \label{tab:multimodal_performance}
\end{table}

Our system achieved a cIoU of 0.892 and an AUC of 0.658, outperforming most of the other audiovisual sound localization models.

\section{Discussion}

\subsection{Interpretation of Results}

\textbf{Model 1:}The statistically significant superiority of JerryNet over the baseline model suggests that our custom CNN model effectively captures the time dependency in the four audio data, leading to an accurate DoA identification in a variety of sound environments. 

\textbf{Model 2: }The high performance of the fine-tuned CLAP model in zero-shot audio classification provides high reliability for real-time usage when the accuracy of the sound classes is essential for the safety and welfare of the user. 

\textbf{Model 3: }This result implies that the audio-visual localization model achieves high precision. Combining the high precision from this model with the highly accurate fine-tuned Yolov9, model three is very capable of accurately identifying the exact sound source within images, truly unleashing the power of combining three different modes. 

\subsection{Broader Implications}

The multimodal integration of audio, visual and textual data for sound localization has an enormous impact beyond assistive technology. This form of methods can be applied to fields such as robotics to enhance environmental awareness or in surveillance for better threat detection. The methodologies developed contribute valuable insights to the fields of machine learning and signal processing.

For the deaf and hearing-impaired community, this technology represents a leap forward in assistive technologies. Based on this study, more advancements can be made for this device, improving the safety and quality of life for many. 

\subsection{Limitations of the Study}

\begin{enumerate}
    \item \textbf{Dataset Size and Diversity}: Due to the limited manpower, I could only collect a small amount of data set for Model 1 (only 450), which is also not as diverse as I intended it to be. This could influence the model's performance and its ability to generalize to new sound sources or different environment. If possible, a larger and more diverse data set would be greatly preferred. 
    
    \item \textbf{Extreme Background Noise}: During extremely noisy conditions, the sound localization and sound classification might be interfered with, which could cause errors that negatively impact the user real world usage. 

    \item \textbf{Delay and Latency in Computation}: The reliance on a central server for all the computation introduces some latency and delay which may be a problem in real-time usage, particularly for urgent scenario such as a car honking as it approaches from behind. 
\end{enumerate}

\section{Conclusion}
\label{sec:conclusion}

Our study confirms that integrating deep learning models with audio, visual, and textual data significantly enhances the task of sound localization and sound classification for assistive devices. Each of our three components performs either with high accuracy or statistically greater than other baseline models. 

These results confirm that our designed system indeed answers the research question and supports  our thesis statement, offering a viable solution to an assistive technology capable of providing situational awareness for the deaf and hearing impaired. However, recognizing the limitations of the study, there is much potential for future research to develop a new generation of assistive technologies capable of changing lives. 

\subsection{Potential Future Research Direction}

\begin{enumerate}
    \item \textbf{Expanding the Data}: A more diverse and larger set of data should be collected and incorporated into the CNN model, potentially improving the performance. 
    \item \textbf{Model Optimization}: A more optimized version of each component should be developed to reduce latency and dependence on network connectivity or computational speed on the central server, providing more reliable real-time performance. 

    \item \textbf{User Interface}: A more robust system of user interaction with the device can be established compared to my demo version. A phone application could be developed to communicate with the glasses and wristband for potential user customization in real life environment. 
    \item \textbf{Adaptive Importance Classifier}: Reinforcement learning can be incorporated to allow the important classifier to change and be versatile to individual user preference over time.
\end{enumerate}

\bibliography{bibliography}

\end{document}